\pdfoutput=1
\documentclass[11pt]{article}

\usepackage[final]{acl}

\usepackage{times}
\usepackage{latexsym}

\usepackage[T1]{fontenc}
\usepackage[utf8]{inputenc}

\usepackage{microtype}

\usepackage{inconsolata}
\usepackage{multirow}
\usepackage{colortbl}
\usepackage{booktabs}
\usepackage{graphicx}
\usepackage{comment}
\usepackage{kotex}

\usepackage{tikz}
\usepackage{pgfplots}
\usepackage{filecontents}
\usepackage{array}
\usepackage{caption}
\usepackage{amsmath}
\usepackage{xcolor}
\usepackage{listings}
\usepackage{float}
\pgfplotsset{compat=1.18}
\newcommand{\indicator}[1]{\mathcal{I}\left(#1\right)}
\newcommand{\abs}[1]{\left\lvert#1\right\rvert}

\title{Dynamic Order Template Prediction for Generative Aspect-Based Sentiment Analysis}

\author{Yonghyun Jun \and Hwanhee Lee\thanks{Corresponding author.} \\
    Department of Artificial Intelligence, Chung-Ang University\\
   \texttt{\{zgold5670, hwanheelee\}@cau.ac.kr}
}

\begin{document}
\maketitle
%\footnotetext{\textsuperscript{$\dagger$}Corresponding author.}

\begin{abstract}
Aspect-based sentiment analysis (ABSA) assesses sentiments towards specific aspects within texts, resulting in detailed sentiment tuples.
Previous ABSA models often used static templates to predict all the elements in the tuples, and these models often failed to accurately capture dependencies between elements.
Multi-view prompting method improves the performance of ABSA by predicting tuples with various templates and then assembling the results.
However, this method suffers from inefficiencies and out-of-distribution errors.
In this paper, we propose a Dynamic Order Template (DOT) method for ABSA, which dynamically creates an order template that contains only the necessary views for each instance. 
Ensuring the diverse and relevant view generation, our proposed method improves F1 scores on ASQP and ACOS datasets while significantly reducing inference time.\footnote{Our implementation is publicly available at \url{https://github.com/imsongpasimin/DOT}}
% This result ensures we success to combine the efficiency of single view with the robustness of multi-view methods.
\end{abstract}

\section{Introduction}
\label{sec:1}
\begin{figure}[t]
  \centering
    \includegraphics[width=0.43\textwidth]{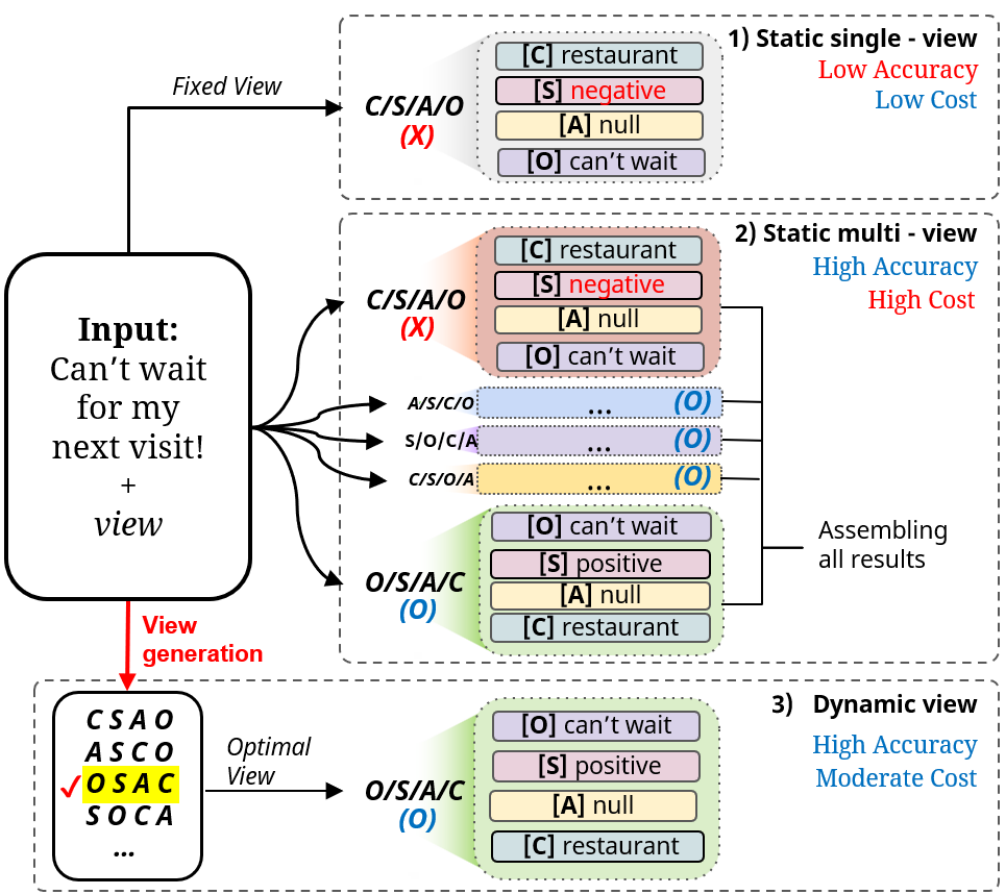}
    %\vspace{-2mm}
    \caption{
    %Comparison of three different order prompting methods. 1) static single-view prediction denotes T5-paraphrase, and 2) static multi-view prediction denotes MvP~\cite{}. Our method is denoted as 3) dynamic-view prediction.
    %Comparison of three different generative ABSA methods. 1) static single-view (T5-paraphrase), 2) static multi-view (MvP), and 3) dynamic-view prediction (ours).
    Comparison of three different generative ABSA methods. 1) static single-view, 2) static multi-view, and 3) dynamic-view prediction (ours).
    %T5-paraphrase reconstructs the original sentence with fixed single order $c\Rightarrow s \Rightarrow a \Rightarrow o$, and extracts tuples from the reconstructed one. Multi-view prompting augments the original sentence with different order prompt templates, and conducts a majority vote within the predicted outputs. The number of templates are set as hyperparameter, so every sentence has the same number of templates. Our method constructs a tuple-wise order prompt templates, and conduct inference based on these template. The number of templates of each sentence is same as its number of tuples.
    }
    \vspace{-6mm}
    \label{fig:1}
\end{figure}

Aspect-based sentiment analysis (ABSA) aims to identify the sentiment of aspects in a given text rather than simply classifying the overall sentiment of the entire text.
ABSA research evolves to generate quadruples consisting of four elements: 1) Aspect ($A$), 2) Category ($C$) for the type of $A$, 3) Opinion ($O$) for $A$, and 4) Sentiment ($S$) for $A$.
Many recent studies, such as T5-paraphrase, tackle this problem using generative models~\citep{zhang2021towards}. 
These approaches usually get review sentences as input and output the span of quadruples in fixed order form, such as "$C$ is $S$ because $A$ is $O$"~\citep{zhang-etal-2021-aspect}. % c->s->a->o.
However, this static single-order template cannot express the dependence between elements as in Figure~\ref{fig:1} due to the autoregressive nature of the transformer~\citep{vaswani2017attention}. Moreover, the model output can heavily depend on the order of generation of each element~\cite{hu-etal-2022-improving-aspect}.

Multi-view prompting~\citep{gou2023mvp} (MvP) deals with this issue by constructing order templates as a channel for "viewing" different perspectives in a sentence. As shown in Figure~\ref{fig:1}, MvP permutes all possible element orders and sorts them based on the entropy of the pre-trained model at the dataset level.
Using this entropy, MvP samples top-k orders and adds these orders as a prompt template. During inference time, MvP conducts majority votes on generated sentiment tuples with various templates. Through this ensemble approach, MvP uses the intuition of solving problems from multiple views in human reasoning and decision~\cite{Stanovich_West_2000}, resulting in enhanced performance.
However, we find that this static multi-view approach of MvP has several drawbacks: \textbf{\textit{1) Inefficiency:}} Even for samples where the answer can be easily found and multiple views are not necessary, this method generates the same number of views, resulting in unnecessary computation that increases the inference time.
\textbf{2) \textit{Limited Transferability}}: 
MvP uses the number of views k as a hyperparameter, applying the same k value across all datasets during training and inference. However, since the optimal number of ensemble models varies according to the data domain, it requires manual adjustment of the k value for each dataset~\cite{shahhosseini2022optimizing}, which hinders the transferability to other datasets. 

To resolve the aforementioned shortcomings, we propose a Dynamic Order Template (DOT) method for ABSA that combines the advantages of both single-view and multi-view approaches.
By prioritizing multiple views based on instance-level entropy, DOT aims to generate only the necessary number of views for each instance during inference.
For an example that contains only one tuple, as in Figure~\ref{fig:1}, DOT dynamically creates only one view as an order template necessary to predict the tuple.
After generating the views, DOT generates tuples using the views inside the order template. 
This phase operates in a multi-view manner, enabling us to retain the benefits of previous multi-view methods. 
Extensive experiments on four widely used sentiment quadruple prediction datasets, derived from ASQP~\cite{pontiki2016semeval, zhang2022survey}, ACOS~\citep{cai-etal-2021-aspect, cai2023memd}, and MEMD-ABSA~\cite{cai2023memd}, demonstrate that our method shows state-of-the-art performance with significantly lower inference time compared to the multi-view approach.
Moreover, we show that our method is robust to domain shift compared to previous methods, resulting in higher transferability.

\section{Method}
\label{sec:2}
\begin{figure}[t]
  \centering
    \includegraphics[width=0.45\textwidth]{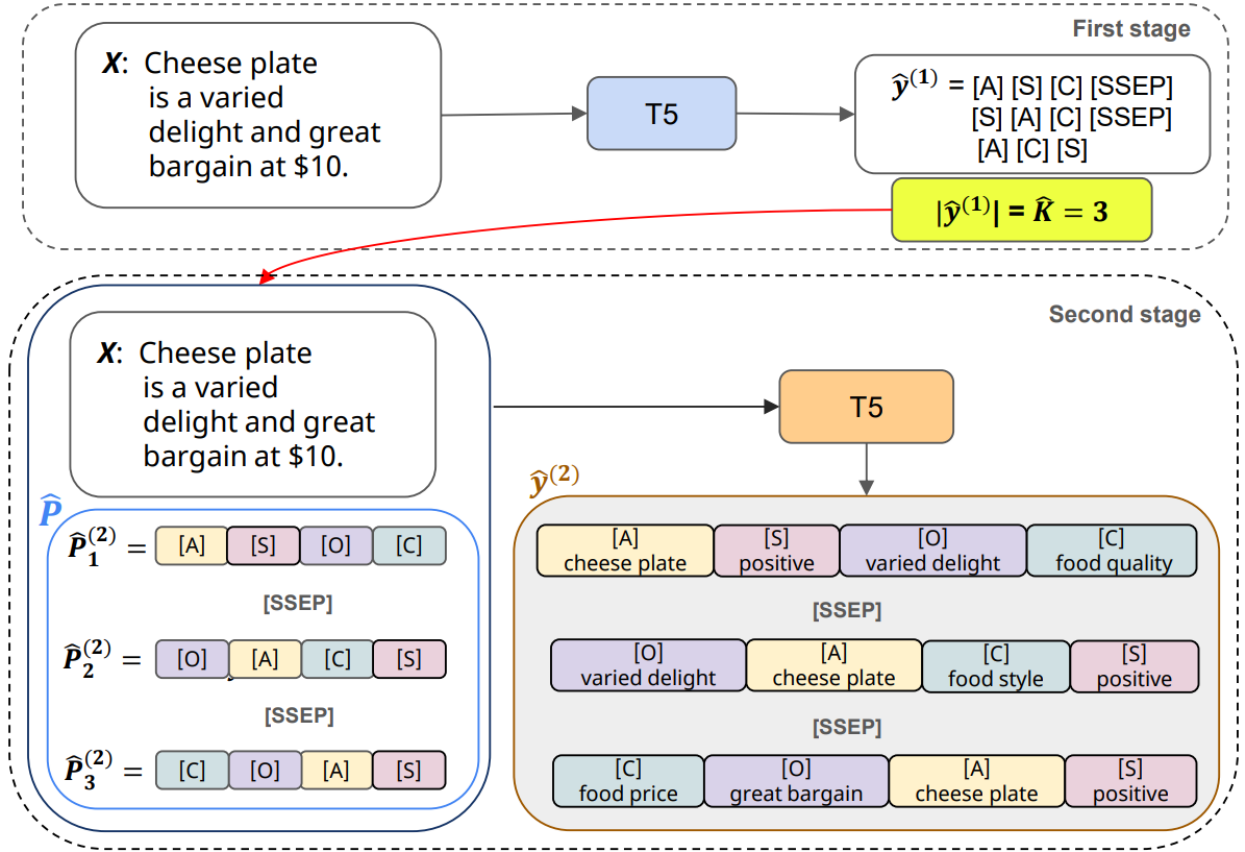}
    \caption{
    Overview of our proposed two stage method. We use two T5 models for each stage: one for generating the initial order template, the other for forming the final order template and generating sentiment tuples.
    }
    \vspace{-6mm}
    \label{fig:2}
\end{figure}

Our proposed Dynamic Order Template (DOT) method is composed of two stages as in Figure~\ref{fig:2}.
The first stage involves generating an initial order template (Sec~\ref{sec:2.1}) to predict the number of tuples.
The second stage involves refining the initial template from stage 1 to produce the final order template and predicting the sentiment tuples based on it (Sec~\ref{sec:2.2}). 
For both stages, we map sentiment tuples ($A$, $C$, $S$, $O$) to marker tokens $[A]$, $[C]$, $[S]$, and $[O]$ respectively. 

Also, for the instances that contain multiple sentiment tuples, we indicate each tuple with the respective tokens and concatenate the targets with \text{[SSEP]} tokens. It is important to note that null or missing values in the datasets are encoded as the literal string "null". During template construction, our method maps this string directly to the designated marker tokens, eliminating the need for any special handling.

\subsection{Stage 1: Generating Order Template}
\label{sec:2.1}
We assume that the number of sentiment tuples $K_i$ in $i^{th}$ instance present for each instance corresponds to the required number of views. Consequently, we are able to divide the complex sentiment tuple generation task into two relatively simpler subtasks: 1) predicting the number of sentiment tuples in the sentence, and 2) generating sentiment tuples with the exact number of first stage predictions. Building separate models specialized for each subtask and enabling collaboration between them was instrumental in achieving significant performance improvements. In this framework, each view is interpreted as the ordered scaffold for generating a single sentiment tuple, as depicted in Figure~\ref{fig:2}.
This allows each prediction order to correspond one-to-one with a sentiment tuple in the second stage. 

We observe that instead of directly using the value of $K_i$ as the target, sampling views corresponding to $K_i$ as the target for the model to generate leads to more accurate prediction of the number of tuples in the given instance (More details are in Appendix~\ref{sec:B}). To establish the view sampling strategy, we start by ranking all possible views generated through permutations through each entropy score, following ~\cite{hu-etal-2022-improving-aspect}.

Specifically, we calculate entropy of each view $v$ in instance-level with vanilla T5 by calculating conditional generation probability as follows:
\begin{equation}%\label{firstloss}
\begin{aligned}
% \mathcal{E}_{i,v} &= p(\boldsymbol{y_{i,v}} | \boldsymbol{x_i}),
\mathcal{E}_{i,v} &= - \sum P(v|{x_i}) \log P(v|{x_i})
\label{eq:1}
\end{aligned}
\end{equation}
Here, $\mathcal{E}_{i,v}$ is the entropy of the total sequence when the $i_{th}$ instance is input into the T5 model and $v$ is the output. At this time, we note that actually utilizing only $A$, $C$, $S$ during the first stage notably facilitates the training process in the second stage. We provide a detailed analysis on excluding $O$ in Appendix~\ref{sec:exclude}.
After computing the entropy, we sort the views by the entropy in ascending order to get the ranked set of views $P^{(1)}_i$. 
And then we sample the top $K_i$ views for each sample and concatenate these views as an order template.

Using the original $i^{th}$ input sentence, we train the T5 model to generate the first-stage target $y^{(1)}_i$ as follows:
%\vspace{-5pt}
\[
y^{(1)}_i = \begin{aligned}[t]
        & P^{(1)}_{i,1} \, \text{[SSEP]} \, P^{(1)}_{i,2} \, \text{[SSEP]} \, \ldots \, P^{(1)}_{i,K_i},
        \end{aligned}
\]
where $P^{(1)}_{i,K_i}$ denotes $K_i^{th}$ view in $P^{(1)}_i$.
We set the loss function to train the T5 model as in Equation~\eqref{eq:loss_s1}, where $\abs{B}$ denotes the batch size of the model. The scaling factor is omitted for simplicity.
\begin{equation}%\label{firstloss}
\begin{aligned}
\mathcal{L}_{1} &= -\sum_{i=1}^{{\abs{B}}} \sum_{t=1}^{T} \log p(\boldsymbol{y^{(1)}_{i,t}} | \boldsymbol{x_i}, \boldsymbol{y^{(1)}_{i,{<t}}})
\label{eq:loss_s1}
\end{aligned}
\end{equation}

\subsection{Stage 2: Sentiment Tuple Generation}
\label{sec:2.2}
In the second stage, the model is trained to generate the sentiment tuple of a given instance using the number of sentiment tuples (i.e. $K_i$). Different from the first stage, we need to generate all elements in sentiment quadruples including $O$ in this stage. 
Hence, we re-rank all views to pick $K_i$ views including $O$ (i.e. ($A$, $C$, $S$, $O$)).

Here, we adopt the same strategy as in the first stage, using entropy to form a ranked set of views, $P^{(2)}_i$. We then sample top $K_i$ views from $P^{(2)}_i$ and add them as an order template prompt $P_i$ to original input sentence.

We design the second stage target $y^{(2)}_i$ by aligning each sentiment tuple with an order template, ensuring that the model learns to generate different tuples for different perspectives. Also, we place the corresponding elements next to each marker token within $P_i$ as follows:

\[
y^{(2)}_i = \begin{aligned}[t]
        & P^{(2)}_{i,1}\otimes tuple_1\, \text{[SSEP]} \, \ldots \, P^{(2)}_{i,K_i}\otimes tuple_{K_i},
        \end{aligned}
\]

where $P^{(2)}_{i,K_i}$ represents $K_i^{th}$ view in $P^{(2)}_i$ and $tuple_{K_i}$ is the $K_i^{th}$ sentiment tuple for given instance. $\otimes$ denotes an interleaved combination between marker tokens and elements. 
Detailed examples for both stages are present in Appendix~\ref{sec:D}.
We design the loss function for training the T5 model in second stage as follows.
\vspace{-7pt}
\begin{equation}%\label{secondloss}
\begin{aligned}
\mathcal{L}_{2} &= -\sum_{i=1}^{{\abs{B}}} \sum_{t=1}^{T} \log p(\boldsymbol{y^{(2)}_{i,t}} | \boldsymbol{x_i}, \boldsymbol{P_i}, \boldsymbol{y^{(2)}_{i,{<t}}})
\label{eq:2}
\end{aligned}
\end{equation}

\subsection{Two-stage inference}
\label{sec:2.3}
During inference time, two stages are conducted sequentially. In the first stage, the model generates the initial order template, denoted as $\hat{y^{(1)}}$. In the second stage, we count the number of generated views from $\hat{y^{(1)}}$ to set $\hat{K}$. Using $\hat{K}$, we sample the top $\hat{K}$ views from the newly ranked set of views and constructs the final order template, referred to as $\hat{P}$. Finally, $\hat{P}$ is directly appended to the inference sentence, enabling the generation of different sentiment tuples for each view in $\hat{P}$. Throughout inference, we employ a constrained‑decoding strategy~\cite{de2020autoregressive} to ensure that the output at each stage conforms to the required format.
The overall two-stage process is described in Figure~\ref{fig:2}.

\section{Experiment}

\begin{table*}[t]
\centering
\setlength{\tabcolsep}{4pt}
\resizebox{0.85\textwidth}{!}{
\begin{tabular}{l||cc|cc|ccccc|c||c}
\toprule
\multirow{2}{*}{\textbf{Methods}} & \multicolumn{2}{c|}{\textbf{ASQP}} & \multicolumn{2}{c|}{\textbf{ACOS}} & \multicolumn{5}{c|}{\textbf{MEMD}} & \multirow{2}{*}{\textbf{Avg}} & \multirow{2}{*}{\textbf{Time(s)}} \\ 
                   & \textbf{R15} & \textbf{R16} & \textbf{Lap} & \textbf{Rest} & \textbf{M-Rest} & \textbf{M-Lap} & \textbf{Books} & \textbf{Clothing} & \textbf{Hotel} & \\
\midrule 
TAS-BERT \small & 34.78 & 43.71 & 27.31 & 33.53 & - & - & - & - & - & - & - \\
Extract-Classify \small & 36.42 & 43.77 & 35.80 & 44.61 & - & - & - & - & - & - & - \\
One-ASQP (large) \small & - & - & 41.56 & 60.69 & - & - & - & - & - & - & - \\
\midrule
Seq2Path \small &   -    &  -  &   42.97  & 58.41 &  - &  - &  - &  - &  - &  - &  - \\  
AugABSA \small & 50.01 & 60.88  &  - &  - &  - &  - &  - &  - &  - &  - &  - \\
SCRAP \small & 49.93 & \textbf{62.48}  &  - &  - &  - &  - &  - &  - &  - &  - &  - \\
Paraphrase \small & 46.93 & 57.93  &  43.51  & \underline{61.16} & 57.38 & 35.07 & 39.30 & 43.00 & 68.79 & 50.34 & 40.63 \\
DLO \small & 48.18  & 59.79  &  43.64 &  59.99 & 57.07 & \underline{35.56} & \underline{42.63} & 43.35 & \textbf{70.27} & 51.16 & 260.74 \\
MvP \small & \underline{51.04} & 60.39  &  \underline{43.92} & \textbf{61.54} & \underline{58.12} & 35.25 & 42.57 & \textbf{43.94} & 69.06 & \underline{51.76} & 2161.81 \\
\midrule
GPT-4o \small & 40.45 & 47.29 & 24.77 & 46.53 & 35.11  & 20.69 & 30.39 & 40.27 & 24.84 & 34.48 & - \\
LLaMa-3.1-8b \small & 37.52 & 47.60  & 40.07 &  54.06 & 38.10  & 31.16 & 28.62 & 32.21 & 44.62 & 39.33 & - \\
Qwen-2.5-7b \small & 29.93 & 39.34 & 12.48 & 33.56 & 25.63  & 24.13 & 17.77 & 18.09 & 38.03 & 26.66 & - \\
Mistral-7b \small & 44.14 & 51.96  & 39.02 & 53.02 & 41.28  & 26.80 & 26.54 & 21.81 & 40.35 & 38.32 & - \\
\midrule
DOT (Ours) & \textbf{51.91} & \underline{61.24}  &  \textbf{44.92} & 59.25 & \textbf{58.25} & \textbf{39.02} & \textbf{43.02} & \underline{43.37} & \underline{69.94} & \textbf{52.28} & 298.17 \\

\bottomrule
\end{tabular}
}
%\vspace{-4mm}
\caption{\label{tab:1}
%F1 scores for ABSA on nine datasets. The best results are in bold and the second best are underlined. We conduct experiments with 5 different seeds and report the average results. Time denotes the averaged inference time. 
F1 scores for ABSA on nine datasets. Best results are in bold, second-best underlined. Results are averaged over five seeds. Time denotes average inference time.
}
\vspace{-4mm}
\end{table*}

\begin{comment}
\begin{tabular}{cccc||c}
\multicolumn{4}{c||}{\textbf{TASD}} & \\
\textbf{R15} & \textbf{R16} & \textbf{L14} & \textbf{R14} & \\
\midrule
58.19 & 70.52 & 60.23 & 69.05 & - \\
63.06 & 71.97 & 61.13 & 72.03 & 62.56 \\
63.89 & 69.23 & \textbf{64.82} & \textbf{75.52} & \underline{65.88} \\
62.95 & 71.79 & 61.46 & 72.39 & 64.26 \\
62.30 & 71.80 & 62.20 & 73.70 & 64.40 \\
64.53 & \underline{72.76} & 63.33 & \underline{74.05} & \textbf{65.89} \\
%64.74 & 70.18 & \underline{65.30} & \textbf{76.30} & \textbf{69.44} & \underline{73.10} & \textbf{63.44}\\
- & - & 62.66 & 73.76 & 65.80 \\
- & 46.51 & 38.12 & - & - \\
- & 65.81 & 49.83 & - & - \\
- & 70.41 & 50 & - & - \\
61.31 & 70.83 & 63.26 & 70.32 & -\\
62.42 & 70.46 & 62.42 & 71.08 & -\\
62.8 & 70.71 & \underline{63.83} & 72.00 & 64.63 \\
\bottomrule
\end{tabular}

\begin{tabular}{cccc||c}
\multicolumn{4}{c||}{\textbf{ASTE}} & \\
\textbf{R14} & \textbf{R15} & \textbf{R16} & \multirow{-2}{*}{\textbf{AVG}} \\
\midrule
69.05 & - & - & - \\
72.03 & 62.56 & 71.70 & 61.20 \\
\textbf{75.52} & \underline{65.88} & 72.87 & - \\
72.39 & 64.26 & {73.03} & \underline{61.75} \\
73.70 & 64.40 &  69.90 & - \\
\underline{74.05} & \textbf{65.89} & \textbf{73.48} & \textbf{63.09}\\
%76.30  & \textbf{69.44} & \underline{73.10} & \textbf{63.44}\\
73.76 & 65.80 & {74.23} & - \\
- & - & - & - \\
- & - & - & - \\
- & - & - & - \\
70.32 & 61.47 & 69.18 & -\\
71.08 & 62.34 & 69.48 & -\\
72.00 & 64.63 & 71.61 & -\\
\bottomrule
\end{tabular}
\end{comment}

\subsection{Benchmark Datasets} 
\label{sec:3.1}
We adopt two widely used ABSA datasets: ASQP and ACOS, where the task is to predict sentiment quadruples. 
For ASQP task, we use rest15 (R15) and rest16 (R16) datasets released from~\cite{pontiki2016semeval, zhang2022survey}. For ACOS task, we use laptop16(Lap) and rest16(Rest) datasets constructed by~\cite{cai-etal-2021-aspect, pontiki2016semeval}. Also, we adopt additional ACOS benchmarks from MEMD datasets (Restaurant, Laptop, Books, Clothing, Hotel)~\cite{xu2023measuring} which use a different source from the previous datasets. We refer to the Restaurant and Laptop datasets in MEMD as M-Rest and M-Laptop, respectively, for the sake of clarity.

\subsection{Baselines}
\label{sec:3.2}
We benchmark our approach against a suite of extraction‑ and generation‑based baselines. For a subset of baselines, we confine evaluation to the four datasets—R15, R16, Laptop, and Restaurant—because these methods were originally optimized for ASQP or ACOS tasks. Other baselines are assessed on their full applicable dataset range. This selective restriction ensures a fair comparison. The complete list of baselines is as follows:
\textit{TAS-BERT}~\cite{wan2020target} jointly extracts and detects sentimental tuples.
\textit{Extract-Classify}~\cite{cai2021aspect} divide the task into two stages: extraction and classification.
\textit{One-ASQP (large)}~\cite{zhou2023unified} identify the aspect-opinion-sentiment (AOS) triplets simultaneously.
\textit{Seq2Path}~\cite{mao-etal-2022-seq2path} generates sentiment tuples as multiple paths of a tree, and automatically selects a valid one.
\textit{AugABSA}~\cite{wang2023generative} generates a original text based on augmented sentiment quadruples.
\textit{SCRAP}~\cite{kim2024self} optimizes the model to generate extract-then-assign reasonings and the corresponding sentiment quadruplets in sequence.
\textit{Paraphrase}~\cite{zhang-etal-2021-aspect}\: formulates a paraphrase generation process for ABSA with a single fixed order.
\textit{DLO}~\cite{hu-etal-2022-improving-aspect}\: augments data via the multiple order templates.
\textit{MvP} aggregates sentiment tuples generated from multiple orders of prompts via ensembling.
Also, we benchmark popular LLMs such as GPT-4o~\cite{hurst2024gpt}, LLaMa-3.1-8b~\cite{dubey2024llama}, Qwen-2.5-7b~\cite{yang2025qwen3}, and Mistral-7b~\cite{jiang2023mistral}.

Detailed setups for LLMs are in Appendix~\ref{sec:llm}.

\subsection{Implementation Details}
\label{sec:3.3}
We utilize the pre-trained T5-base~\citep{raffel2020exploring} model as the backbone for the first stage. We also use the model trained in the first stage as the backbone for the second stage, allowing us to leverage a tuned initial point for the ABSA dataset to have the regularization effect inspired by~\cite{fu2023effectiveness}. 

Additionally, we observe that the label of the datasets (i.e. sentiment tuples) irregularly contains stop words. Thus, we eliminate irregularities in tuples through stop-word filtering in the second stage. We provide a detailed analysis and filtering process in Appendix~\ref{sec:det_exp}.

\subsection{Results}
\label{sec:3.4}
\paragraph{Performance Comparison} We use the F1 score, which is a standard metric for ABSA, to measure the performance of the systems. 

Table~\ref{tab:1} demonstrates that our DOT framework achieves state‑of‑the‑art results, ranking first on five of the nine benchmarks and second on three others. To further elaborate, a significant factor contributing to the enhanced performance is our approach of dividing the complex sentiment quadruple generation task into two subtasks, each handled by dedicated models. As a result, our method achieves better performance compared to a single model handling multi-view processing alone. Crucially, the effectiveness of this two-stage approach depends on the accuracy of the first task, which we comprehensively presented in Appendix~\ref{sec:B}. However, its performance is slightly lower on the Rest and Clothing datasets, which we analyze in Section~\ref{sec:implicit}.

\paragraph{Inference time} We also measure inference time using T5-base model for all baselines. We check inference time for each dataset, and average them.

As in Table~\ref{tab:1}, we dramatically reduce inference time particularly compared to the multi-view methods such as MvP~\cite{gou2023mvp}, by predicting solely the necessary number of views for each sample.

On the other hand, single view inferences~\cite{zhang-etal-2021-aspect, hu-etal-2022-improving-aspect} and extraction approaches~\cite{wan2020target, cai2021aspect, zhou2023unified} can be more memory‑ and time‑efficient\cite{zhou2023unified}, they generally exhibit inferior performance compared to generative methods.
We provide more details on the inference time in Appendix~\ref{sec:C}.

\paragraph{Transferability}
\label{sec:ood}

\begin{table}[h]
\centering
\setlength{\tabcolsep}{4pt}
\resizebox{0.43\textwidth}{!}{
\begin{tabular}{l|cc||cc}
\toprule
\textbf{Train} & \multicolumn{2}{c||}{\textit{SemEval}} & \multicolumn{2}{c}{\textit{Yelp}} \\ 
\midrule
\textbf{Test} & \textit{SemEval} & \textit{Yelp} & \textit{Yelp} & \textit{SemEval} \\
\midrule 
Paraphrase & 52.38 & 38.52\textsubscript{\textcolor{blue}{(-11.86)}} & 57.38 & 44.88\textsubscript{\textcolor{blue}{(-12.50)}} \\
MvP\(_{3}\)& 55.62 & 34.42\textsubscript{\textcolor{blue}{(-21.20)}} & 57.27 & 41.72\textsubscript{\textcolor{blue}{(-15.55)}} \\ 
MvP\(_{9}\) & 56.89 & 35.02\textsubscript{\textcolor{blue}{(-21.87)}} & 56.98 & 42.52\textsubscript{\textcolor{blue}{(-14.46)}} \\ 
MvP\(_{15}\) & \textbf{57.66} & 35.21\textsubscript{\textcolor{blue}{(-21.45)}} & 58.12 & 41.94\textsubscript{\textcolor{blue}{(-16.18)}} \\
DOT & 57.47 & \textbf{39.88}\textsubscript{\textcolor{blue}{(-17.59)}} & \textbf{58.25} & \textbf{46.97}\textsubscript{\textcolor{blue}{(-11.28)}} \\
\bottomrule
\end{tabular}
}
%\vspace{-1mm}
\caption{ 
Cross-dataset evaluation results for validating the effect of domain shift.}
\label{tab:cross}
%\vspace{-2mm}
\end{table}

To examine the transferability of each model, we conduct an in-depth experiment on cross-dataset evaluation. 

We group the datasets into two sources: SemEval~\cite{pontiki2016semeval} (R15, R16, Rest) and Yelp (M-Rest), and assess performance by training on one group and testing on the other in a zero-shot setting. For the MvP model, we vary the number of views (3, 9, and 15) to evaluate sensitivity in static multi-view methods, while T5-paraphrase uses a static single order. As shown in Table~\ref{tab:cross}, our model significantly outperforms the baselines in cross-dataset evaluation. Although T5-paraphrase suffers a smaller performance drop, it still falls behind our method, and MvP shows notable degradation regardless of view count. These results demonstrate that our model effectively identifies optimal views even for out-of-domain datasets.

\paragraph{Performance Across Model Scales}
Beyond using T5‑base as the sole backbone, we extend our experiments to examine how performance varies with model scale, specifically testing T5‑small and T5‑large. In this setup, we employ the same backbone model for both stages.
\begin{figure}[!htbp]
  \centering
    \includegraphics[width=0.48\textwidth]{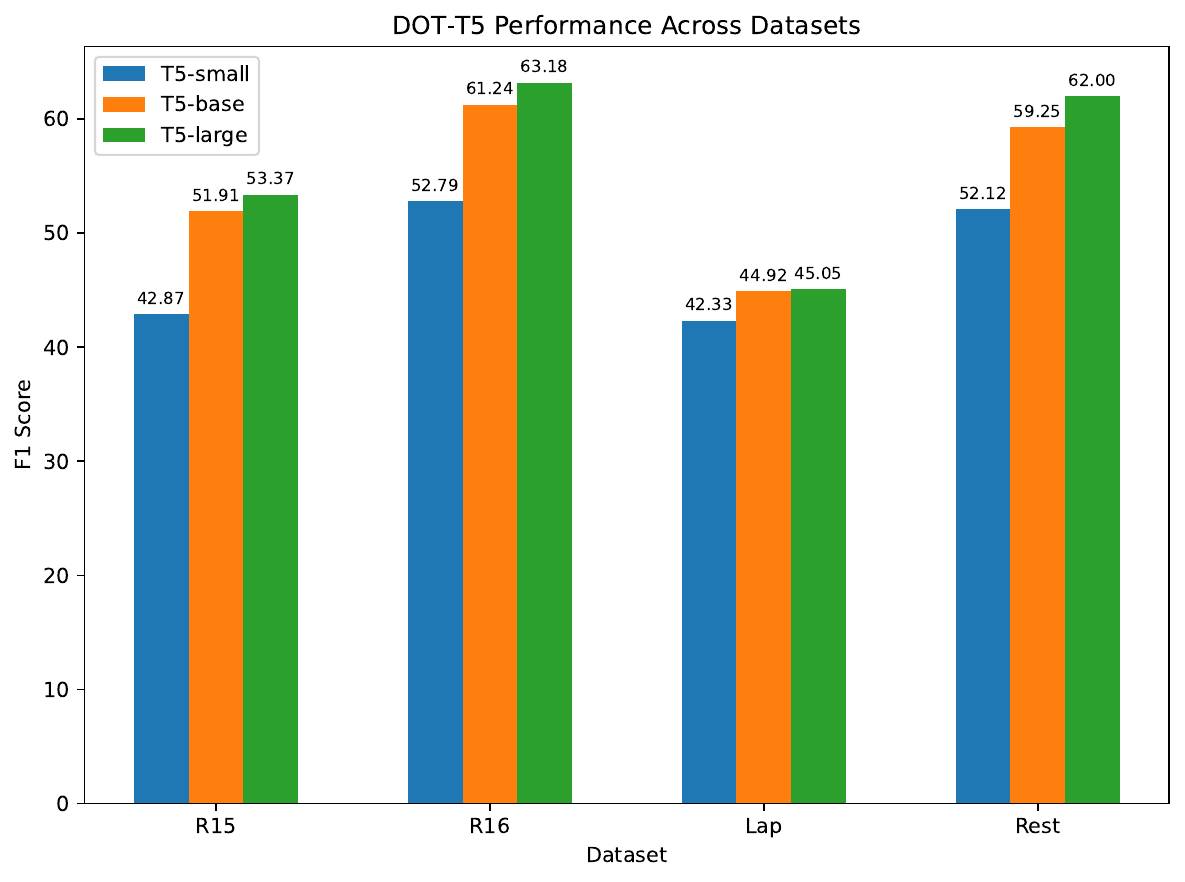}
    \caption{
    Bar chart illustrating F1 performance variations across T5-small, T5-base, and T5-large backbones on R15, R16, Laptop, and Rest benchmarks.
    }
    \label{fig:scale}
\end{figure}
As shown in Figure~\ref{fig:scale}, performance consistently improves across all benchmarks as the backbone scale increases, supporting the scaling law~\cite{kaplan2020scaling} showing larger models yield better accuracy. Nonetheless, these gains do not fully compensate for the inefficiency introduced by loading larger models and performing multiple inference steps. Considering the trade‑off between accuracy and computational cost, T5‑base emerges as the most balanced choice.

\paragraph{Ablation study}
To further investigate the effectiveness of each component of our framework, we conduct an ablation study and present the results in Table~\ref{tab:ablation}.
Firstly, we record the results without stop-word filtering.
Also, we unify the two stages into one, directly generating multiple order templates and tuples without including order prompting.

Additionally, we evaluate the results of sampling the views randomly, checking whether the entropy score is valid. 
Lastly, we exclude the multi-view approach by training and testing our model using only the view with the lowest entropy for each instance as the order template. 
We perform an ablation study by excluding one or more of the four components of our method mentioned earlier.
By observing the gaps between these variants with the original model, we verify the effectiveness of each component of our method.

\begin{table}[t]
    \centering
    \label{tab:3}
    \small
    \resizebox{0.48\textwidth}{!}{
    \begin{tabular}{lcc}
        \toprule
        \textbf{Model Configuration} & \textbf{Average F1} \\
        \midrule
        Full Model & 54.33 & \\
        \midrule
        w/o filtering & 53.31 \textsubscript{\textcolor{blue}{(-1.02)}} \\
        w/o stage division & 52.73 \textsubscript{\textcolor{blue}{(-1.60)}} \\
        w/o entropy score & 52.53 \textsubscript{\textcolor{blue}{(-1.80)}} \\
        w/o multi view & 52.31 \textsubscript{\textcolor{blue}{(-2.02)}} \\
        w/o stage division, entropy score & 50.04 \textsubscript{\textcolor{blue}{(-4.29)}} \\
        w/o filtering, stage division, entropy score & 45.80 \textsubscript{\textcolor{blue}{(-8.53)}} \\
        \bottomrule
    \end{tabular}
    }
    %\vspace{-1mm}
    \caption{Ablation study for the proposed method, which shows the average F1 across ASQP and ACOS.}
\label{tab:ablation}
\vspace{-3mm}
\end{table}

\section{Conclusion}
We propose Dynamic Order Template (DOT) method for generative ABSA, addressing inefficiencies and out-of-distribution errors. 
Experiments on nine datasets demonstrate that DOT achieves state-of-the-art performance with reduced inference time, effectively balancing the strengths of previous single and multi-view approaches for ABSA.

\section*{Limitation}
Our DOT method is highly efficient and powerful, yet it still has several limitations.
DOT method consists of two stages: view generation and tuple generation. We train separate models for each task, and these two models perform inference sequentially. This form is not end-to-end, so it is disadvantageous in terms of training time and memory. 

Also, since we directly connect first stage and second stage, if any errors occur, the errors may propagate and magnify as it moves to the subsequent stage. It results in relatively large standard deviation for different seeds as reported in Table~\ref{tab:sd}.
However, by splitting the task of 'predicting the appropriate number of tuples' into two sub-tasks—'predicting the appropriate number of tuples' and 'accurately predicting the tuples'—it becomes significantly easier to achieve accurate results in both areas, thereby enhancing overall performance in our work.

Finally, we define the number of necessary views as the number of sentiment tuples for simplicity and efficiency. A more complex yet refined method for determining the necessary number of views could be further explored in future research.

\section*{Ethics Statement}
This study utilizes the various datasets for aspect-based sentiment analysis, which are accessible online. Additionally, we have properly cited all the papers and sources referenced in our paper. We plan to release the pre-trained model and the code for training the proposed system.

\section*{Acknowledgement}
This work was supported by the Institute of Information \& Communications Technology Planning \& Evaluation (IITP) grant funded by the Korea government (MSIT) [RS-2021-II211341, Artificial Intelligence Graduate School Program (Chung-Ang University)].

\bibliography{acl_latex}

\begin{thebibliography}{35}
\expandafter\ifx\csname natexlab\endcsname\relax\def\natexlab#1{#1}\fi

\bibitem[{Brown et~al.(2020)Brown, Mann, Ryder, Subbiah, Kaplan, Dhariwal, Neelakantan, Shyam, Sastry, Askell et~al.}]{brown2020language}
Tom Brown, Benjamin Mann, Nick Ryder, Melanie Subbiah, Jared~D Kaplan, Prafulla Dhariwal, Arvind Neelakantan, Pranav Shyam, Girish Sastry, Amanda Askell, et~al. 2020.
\newblock Language models are few-shot learners.
\newblock \emph{Advances in neural information processing systems}, 33:1877--1901.

\bibitem[{Cai et~al.(2023)Cai, Song, Wang, Xie, Zhao, Li, Wu, Liu, Yu, and Xia}]{cai2023memd}
Hongjie Cai, Nan Song, Zengzhi Wang, Qiming Xie, Qiankun Zhao, Ke~Li, Siwei Wu, Shijie Liu, Jianfei Yu, and Rui Xia. 2023.
\newblock Memd-absa: a multi-element multi-domain dataset for aspect-based sentiment analysis.
\newblock \emph{arXiv preprint arXiv:2306.16956}.

\bibitem[{Cai et~al.(2021{\natexlab{a}})Cai, Xia, and Yu}]{cai-etal-2021-aspect}
Hongjie Cai, Rui Xia, and Jianfei Yu. 2021{\natexlab{a}}.
\newblock Aspect-category-opinion-sentiment quadruple extraction with implicit aspects and opinions.
\newblock In \emph{Proceedings of the 59th Annual Meeting of the Association for Computational Linguistics and the 11th International Joint Conference on Natural Language Processing (Volume 1: Long Papers)}, pages 340--350.

\bibitem[{Cai et~al.(2021{\natexlab{b}})Cai, Xia, and Yu}]{cai2021aspect}
Hongjie Cai, Rui Xia, and Jianfei Yu. 2021{\natexlab{b}}.
\newblock Aspect-category-opinion-sentiment quadruple extraction with implicit aspects and opinions.
\newblock In \emph{Proceedings of the 59th Annual Meeting of the Association for Computational Linguistics and the 11th International Joint Conference on Natural Language Processing (Volume 1: Long Papers)}, pages 340--350.

\bibitem[{Chebolu et~al.(2023)Chebolu, Dernoncourt, Lipka, and Solorio}]{chebolu2023review}
Siva Uday~Sampreeth Chebolu, Franck Dernoncourt, Nedim Lipka, and Thamar Solorio. 2023.
\newblock A review of datasets for aspect-based sentiment analysis.
\newblock In \emph{Proceedings of the 13th International Joint Conference on Natural Language Processing and the 3rd Conference of the Asia-Pacific Chapter of the Association for Computational Linguistics (Volume 1: Long Papers)}, pages 611--628.

\bibitem[{De~Cao et~al.(2020)De~Cao, Izacard, Riedel, and Petroni}]{de2020autoregressive}
Nicola De~Cao, Gautier Izacard, Sebastian Riedel, and Fabio Petroni. 2020.
\newblock Autoregressive entity retrieval.
\newblock \emph{arXiv preprint arXiv:2010.00904}.

\bibitem[{Dubey et~al.(2024)Dubey, Jauhri, Pandey, Kadian, Al-Dahle, Letman, Mathur, Schelten, Yang, Fan et~al.}]{dubey2024llama}
Abhimanyu Dubey, Abhinav Jauhri, Abhinav Pandey, Abhishek Kadian, Ahmad Al-Dahle, Aiesha Letman, Akhil Mathur, Alan Schelten, Amy Yang, Angela Fan, et~al. 2024.
\newblock The llama 3 herd of models.
\newblock \emph{arXiv preprint arXiv:2407.21783}.

\bibitem[{Farkiya et~al.(2015)Farkiya, Saini, Sinha, and Desai}]{farkiya2015natural}
Alabhya Farkiya, Prashant Saini, Shubham Sinha, and Sharmishta Desai. 2015.
\newblock Natural language processing using nltk and wordnet.
\newblock \emph{Int. J. Comput. Sci. Inf. Technol}, 6(6):5465--5469.

\bibitem[{Fu et~al.(2023)Fu, Yang, So, Lam, Bing, and Collier}]{fu2023effectiveness}
Zihao Fu, Haoran Yang, Anthony Man-Cho So, Wai Lam, Lidong Bing, and Nigel Collier. 2023.
\newblock On the effectiveness of parameter-efficient fine-tuning.
\newblock In \emph{Proceedings of the AAAI Conference on Artificial Intelligence}, volume~37, pages 12799--12807.

\bibitem[{Gou et~al.(2023)Gou, Guo, and Yang}]{gou2023mvp}
Zhibin Gou, Qingyan Guo, and Yujiu Yang. 2023.
\newblock Mvp: Multi-view prompting improves aspect sentiment tuple prediction.
\newblock \emph{arXiv preprint arXiv:2305.12627}.

\bibitem[{Hu et~al.(2021)Hu, Shen, Wallis, Allen-Zhu, Li, Wang, Wang, and Chen}]{hu2021lora}
Edward~J Hu, Yelong Shen, Phillip Wallis, Zeyuan Allen-Zhu, Yuanzhi Li, Shean Wang, Lu~Wang, and Weizhu Chen. 2021.
\newblock Lora: Low-rank adaptation of large language models.
\newblock \emph{arXiv preprint arXiv:2106.09685}.

\bibitem[{Hu et~al.(2022)Hu, Wu, Gao, Bai, and Zhao}]{hu-etal-2022-improving-aspect}
Mengting Hu, Yike Wu, Hang Gao, Yinhao Bai, and Shiwan Zhao. 2022.
\newblock Improving aspect sentiment quad prediction via template-order data augmentation.
\newblock In \emph{Proceedings of the 2022 Conference on Empirical Methods in Natural Language Processing}, pages 7889--7900.

\bibitem[{Hurst et~al.(2024)Hurst, Lerer, Goucher, Perelman, Ramesh, Clark, Ostrow, Welihinda, Hayes, Radford et~al.}]{hurst2024gpt}
Aaron Hurst, Adam Lerer, Adam~P Goucher, Adam Perelman, Aditya Ramesh, Aidan Clark, AJ~Ostrow, Akila Welihinda, Alan Hayes, Alec Radford, et~al. 2024.
\newblock Gpt-4o system card.
\newblock \emph{arXiv preprint arXiv:2410.21276}.

\bibitem[{Jiang et~al.(2023)Jiang, Sablayrolles, Mensch, Bamford, Chaplot, Casas, Bressand, Lengyel, Lample, Saulnier et~al.}]{jiang2023mistral}
Albert~Q Jiang, Alexandre Sablayrolles, Arthur Mensch, Chris Bamford, Devendra~Singh Chaplot, Diego de~las Casas, Florian Bressand, Gianna Lengyel, Guillaume Lample, Lucile Saulnier, et~al. 2023.
\newblock Mistral 7b.
\newblock \emph{arXiv preprint arXiv:2310.06825}.

\bibitem[{Kaplan et~al.(2020)Kaplan, McCandlish, Henighan, Brown, Chess, Child, Gray, Radford, Wu, and Amodei}]{kaplan2020scaling}
Jared Kaplan, Sam McCandlish, Tom Henighan, Tom~B Brown, Benjamin Chess, Rewon Child, Scott Gray, Alec Radford, Jeffrey Wu, and Dario Amodei. 2020.
\newblock Scaling laws for neural language models.
\newblock \emph{arXiv preprint arXiv:2001.08361}.

\bibitem[{Kim et~al.(2024)Kim, Heo, Seo, Kang, Yeo, and Lee}]{kim2024self}
Jieyong Kim, Ryang Heo, Yongsik Seo, SeongKu Kang, Jinyoung Yeo, and Dongha Lee. 2024.
\newblock Self-consistent reasoning-based aspect-sentiment quad prediction with extract-then-assign strategy.
\newblock \emph{arXiv preprint arXiv:2403.00354}.

\bibitem[{Lewis(2019)}]{lewis2019bart}
M~Lewis. 2019.
\newblock Bart: Denoising sequence-to-sequence pre-training for natural language generation, translation, and comprehension.
\newblock \emph{arXiv preprint arXiv:1910.13461}.

\bibitem[{Liu et~al.(2019)Liu, Ott, Goyal, Du, Joshi, Chen, Levy, Lewis, Zettlemoyer, and Stoyanov}]{liu2019roberta}
Yinhan Liu, Myle Ott, Naman Goyal, Jingfei Du, Mandar Joshi, Danqi Chen, Omer Levy, Mike Lewis, Luke Zettlemoyer, and Veselin Stoyanov. 2019.
\newblock Roberta: A robustly optimized bert pretraining approach.
\newblock \emph{arXiv preprint arXiv:1907.11692}.

\bibitem[{Loshchilov and Hutter(2017)}]{loshchilov2017decoupled}
Ilya Loshchilov and Frank Hutter. 2017.
\newblock Decoupled weight decay regularization.
\newblock \emph{arXiv preprint arXiv:1711.05101}.

\bibitem[{Mao et~al.(2022)Mao, Shen, Yang, Zhu, and Cai}]{mao-etal-2022-seq2path}
Yue Mao, Yi~Shen, Jingchao Yang, Xiaoying Zhu, and Longjun Cai. 2022.
\newblock Seq2path: Generating sentiment tuples as paths of a tree.
\newblock In \emph{Findings of the Association for Computational Linguistics: ACL 2022}, pages 2215--2225.

\bibitem[{Pontiki et~al.(2016)Pontiki, Galanis, Papageorgiou, Androutsopoulos, Manandhar, Mohammad, Al-Ayyoub, Zhao, Qin, De~Clercq et~al.}]{pontiki2016semeval}
Maria Pontiki, Dimitrios Galanis, Harris Papageorgiou, Ion Androutsopoulos, Suresh Manandhar, AL-Smadi Mohammad, Mahmoud Al-Ayyoub, Yanyan Zhao, Bing Qin, Orphee De~Clercq, et~al. 2016.
\newblock Semeval-2016 task 5: Aspect based sentiment analysis.
\newblock In \emph{Proceedings of the 10th International Workshop on Semantic Evaluation (SemEval-2016)}, pages 19--30.

\bibitem[{Raffel et~al.(2020)Raffel, Shazeer, Roberts, Lee, Narang, Matena, Zhou, Li, and Liu}]{raffel2020exploring}
Colin Raffel, Noam Shazeer, Adam Roberts, Katherine Lee, Sharan Narang, Michael Matena, Yanqi Zhou, Wei Li, and Peter~J Liu. 2020.
\newblock Exploring the limits of transfer learning with a unified text-to-text transformer.
\newblock \emph{Journal of machine learning research}, 21(140):1--67.

\bibitem[{Shahhosseini et~al.(2022)Shahhosseini, Hu, and Pham}]{shahhosseini2022optimizing}
Mohsen Shahhosseini, Guiping Hu, and Hieu Pham. 2022.
\newblock Optimizing ensemble weights and hyperparameters of machine learning models for regression problems.
\newblock \emph{Machine Learning with Applications}, 7:100251.

\bibitem[{Solnyshkina et~al.(2017)Solnyshkina, Zamaletdinov, Gorodetskaya, and Gabitov}]{solnyshkina2017evaluating}
Marina Solnyshkina, Radif Zamaletdinov, Ludmila Gorodetskaya, and Azat Gabitov. 2017.
\newblock Evaluating text complexity and flesch-kincaid grade level.
\newblock \emph{Journal of social studies education research}, 8(3):238--248.

\bibitem[{Stanovich and West(2000)}]{Stanovich_West_2000}
Keith~E. Stanovich and Richard~F. West. 2000.
\newblock \href {https://doi.org/10.1017/S0140525X00623439} {Advancing the rationality debate}.
\newblock \emph{Behavioral and Brain Sciences}, 23(5):701–717.

\bibitem[{Vaswani et~al.(2017)Vaswani, Shazeer, Parmar, Uszkoreit, Jones, Gomez, Kaiser, and Polosukhin}]{vaswani2017attention}
Ashish Vaswani, Noam Shazeer, Niki Parmar, Jakob Uszkoreit, Llion Jones, Aidan~N Gomez, {\L}ukasz Kaiser, and Illia Polosukhin. 2017.
\newblock Attention is all you need.
\newblock \emph{Advances in neural information processing systems}, 30.

\bibitem[{Wan et~al.(2020)Wan, Yang, Du, Liu, Qi, and Pan}]{wan2020target}
Hai Wan, Yufei Yang, Jianfeng Du, Yanan Liu, Kunxun Qi, and Jeff~Z Pan. 2020.
\newblock Target-aspect-sentiment joint detection for aspect-based sentiment analysis.
\newblock In \emph{Proceedings of the AAAI conference on artificial intelligence}, volume~34, pages 9122--9129.

\bibitem[{Wang et~al.(2023)Wang, Jiang, Ma, Liu, and Okazaki}]{wang2023generative}
An~Wang, Junfeng Jiang, Youmi Ma, Ao~Liu, and Naoaki Okazaki. 2023.
\newblock Generative data augmentation for aspect sentiment quad prediction.
\newblock In \emph{Proceedings of the 12th Joint Conference on Lexical and Computational Semantics (* SEM 2023)}, pages 128--140.

\bibitem[{Wei et~al.(2021)Wei, Bosma, Zhao, Guu, Yu, Lester, Du, Dai, and Le}]{wei2021finetuned}
Jason Wei, Maarten Bosma, Vincent~Y Zhao, Kelvin Guu, Adams~Wei Yu, Brian Lester, Nan Du, Andrew~M Dai, and Quoc~V Le. 2021.
\newblock Finetuned language models are zero-shot learners.
\newblock \emph{arXiv preprint arXiv:2109.01652}.

\bibitem[{Xu et~al.(2023)Xu, Yang, Wu, Chen, Zhao, and Dai}]{xu2023measuring}
Ting Xu, Huiyun Yang, Zhen Wu, Jiaze Chen, Fei Zhao, and Xinyu Dai. 2023.
\newblock Measuring your aste models in the wild: A diversified multi-domain dataset for aspect sentiment triplet extraction.
\newblock In \emph{Findings of the Association for Computational Linguistics: ACL 2023}, pages 2837--2853.

\bibitem[{Yang et~al.(2025)Yang, Li, Yang, Zhang, Hui, Zheng, Yu, Gao, Huang, Lv et~al.}]{yang2025qwen3}
An~Yang, Anfeng Li, Baosong Yang, Beichen Zhang, Binyuan Hui, Bo~Zheng, Bowen Yu, Chang Gao, Chengen Huang, Chenxu Lv, et~al. 2025.
\newblock Qwen3 technical report.
\newblock \emph{arXiv preprint arXiv:2505.09388}.

\bibitem[{Zhang et~al.(2021{\natexlab{a}})Zhang, Deng, Li, Bing, and Lam}]{zhang-etal-2021-aspect}
Wenxuan Zhang, Yang Deng, Xin Li, Lidong Bing, and Wai Lam. 2021{\natexlab{a}}.
\newblock Aspect-based sentiment analysis in question answering forums.
\newblock In \emph{Findings of the Association for Computational Linguistics: EMNLP 2021}, pages 4582--4591.

\bibitem[{Zhang et~al.(2021{\natexlab{b}})Zhang, Li, Deng, Bing, and Lam}]{zhang2021towards}
Wenxuan Zhang, Xin Li, Yang Deng, Lidong Bing, and Wai Lam. 2021{\natexlab{b}}.
\newblock Towards generative aspect-based sentiment analysis.
\newblock In \emph{Proceedings of the 59th Annual Meeting of the Association for Computational Linguistics and the 11th International Joint Conference on Natural Language Processing (Volume 2: Short Papers)}, pages 504--510.

\bibitem[{Zhang et~al.(2022)Zhang, Li, Deng, Bing, and Lam}]{zhang2022survey}
Wenxuan Zhang, Xin Li, Yang Deng, Lidong Bing, and Wai Lam. 2022.
\newblock A survey on aspect-based sentiment analysis: Tasks, methods, and challenges.
\newblock \emph{IEEE Transactions on Knowledge and Data Engineering}.

\bibitem[{Zhou et~al.(2023)Zhou, Yang, He, Mou, and Yang}]{zhou2023unified}
Junxian Zhou, Haiqin Yang, Yuxuan He, Hao Mou, and JunBo Yang. 2023.
\newblock A unified one-step solution for aspect sentiment quad prediction.
\newblock \emph{arXiv preprint arXiv:2306.04152}.

\end{thebibliography}

\clearpage

\appendix

\section{Detailed Experimental Setup}
\label{sec:det_exp}
We use AdamW optimizer~\cite{loshchilov2017decoupled} with a learning rate of 1e-4 for training two T5 models. We set the batch size to 16 for training and 24 for inference. We train the first stage model for 30 epochs, and train 40 epochs for the second stage.

\begin{figure}[!h]
  \centering
    \includegraphics[width=0.48\textwidth]{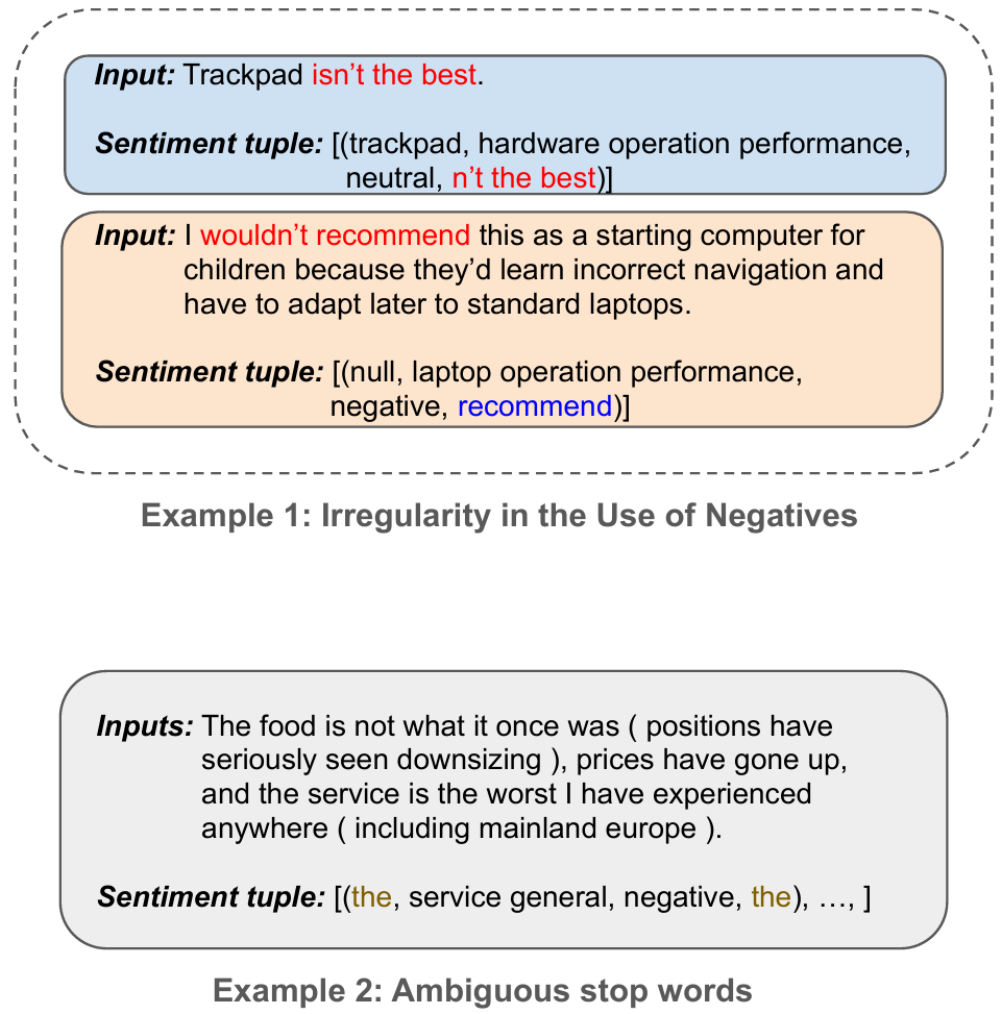}
    \caption{
    Two examples of irregularity of stop words. Note that these examples are the not all of the stop-word problems.
    }
    %\vspace{-3mm}
    \label{fig:stop_words}
\end{figure}

We also note that the dataset labels (i.e., the sentiment tuples) sporadically include stop words. For example, as in the first example of Figure~\ref{fig:stop_words}, the inclusion of negations in the opinion terms is inconsistent. Also, as in the second example, element tuples sometimes contain ambiguous and meaningless stop words as elements. As a result, the fine-tuned model sometimes generates sentiment tuples containing stop words irregularly. It can yield critical performance degradation, even though they don't affect the meaning of the sentiment elements. To resolve the problem of stop words, we filter these stop words using nltk package\citep{farkiya2015natural} for both generated results and dataset labels. We use four RTX 4090 GPUs to train and evaluate all of the models.

\section{Case Study}
\label{sec:case_study}

\begin{figure*}[htbp]
  \centering
    \includegraphics[width=0.95\textwidth]{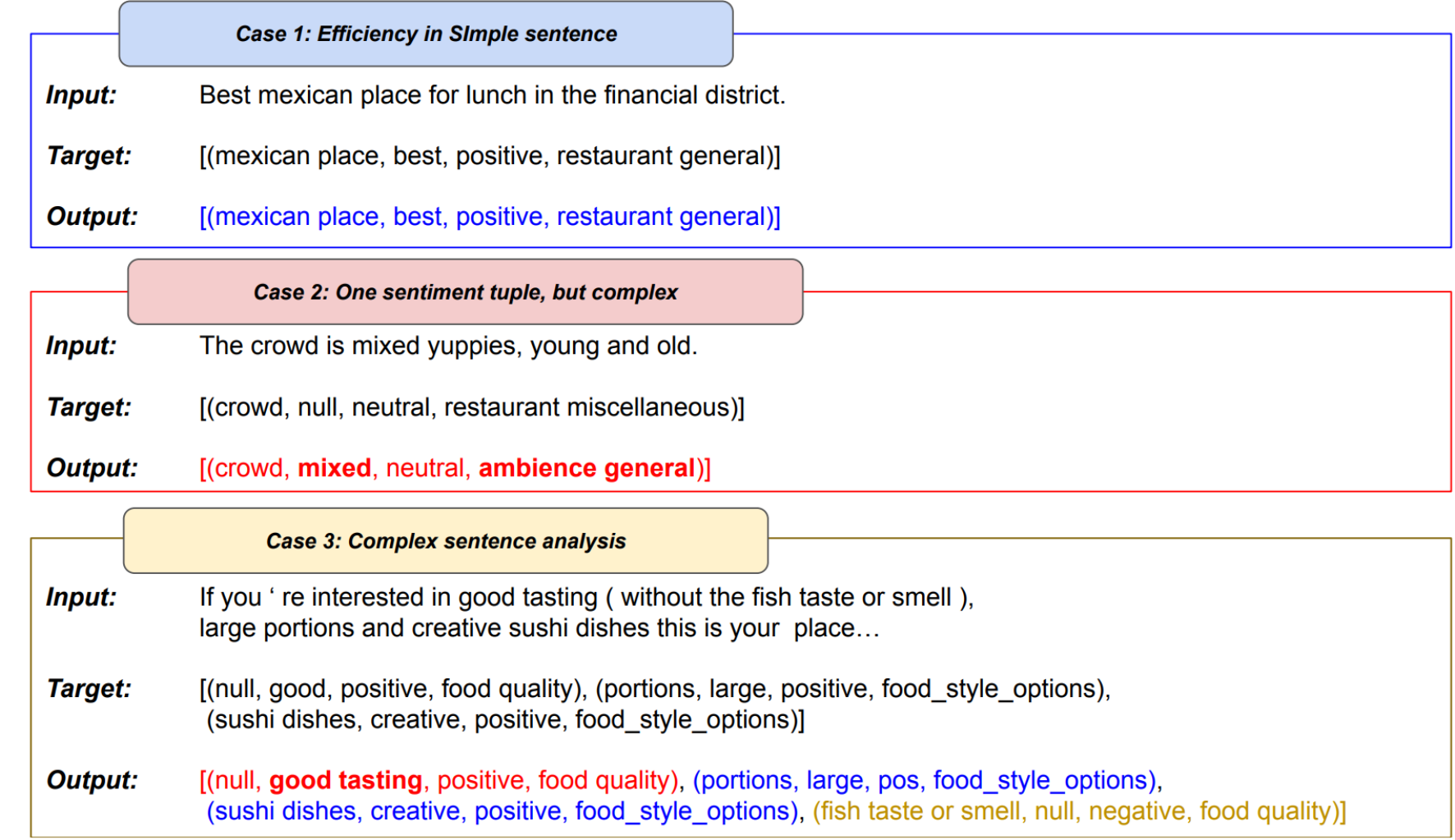}
    \caption{
    Case study for the three main types of results. Blue one denotes correct, red one denotes incorrect, and the yellow one denotes irrelevant.
    }
    \label{fig:6}
\end{figure*}
We conduct a case study and analyze the properties of the outputs generated by the proposed method. As depicted in Figure~\ref{fig:6}, we classify the output results into three main cases.

The first case involves sentences that do not require multiple views for accurate prediction. For these sentences, our model succeeds in making efficient predictions using only a single view. We observe that this case is the most common type in our study, significantly contributing to the model's efficiency.

The second shows an example that predicts requires fewer views, but the example actually requires more views. Our analysis reveals that such cases frequently occurs with implicit $O$. As shown in Table~\ref{tab:1}, this suggests that our model's performance might lag behind other baselines on the ACOS Rest16 dataset, which contains many samples with implicit $A$ and $O$. Additionally, the model struggles with predicting infrequent $C$ in the training set. Incorporating the concept of self-information and defining the necessary number of views based on the 'amount of information in a sample' could effectively address this issue.

The final case involves cases with multiple sentiment tuples and longer lengths. We explain that errors in this scenario stem from two main reasons. Firstly, longer sentences include extended phrases that modify $A$ or $O$. Including all these modifiers as elements often leads to errors, a common problem across different models that requires an alternative solution. Secondly, errors occur when the number of tuples is incorrectly predicted in the first stage. If the predicted number of tuples is insufficient, some target sentiment tuples might be overlooked. Conversely, overestimation leads to the extraction of irrelevant aspects, as depicted in the Figure~\ref{fig:6}. However, we optimize the first stage to reduce tuple count errors, which helped mitigate performance drops by minimizing incorrectly generated or overlooked tuples.

\section{Depth Analysis on First Stage}
\label{sec:B}
\subsection{Accuracy on the Number of Views}
We assess the accuracy of predicting the value of $\hat{K}$ and present the results in Table~\ref{tab:first}. We evaluate the output by comparing it to the number of labeled sentiment tuples using RMSE and accuracy. 
%This measurement can approximately evaluate the performance on catching the number of views in the given sentences.
%\subsection{Baselines}
We carefully implement the first stage baselines to compare our method properly as follows: 
\textit{Random}: We find that the number of sentiment tuples in the training dataset is mostly in the range of 1 to 6. 
%The distribution of the number of tuples is reported in Appendix.
For each inference, we randomly sample one of the 6 numbers and compare it with our first stage result.
\textit{Majority}: We also reveal that about 60 percent of labels consist of a single tuple. We construct a baseline that predicts only 1 for the number of tuples, to check whether our model has the ability to predict the number of sentiment tuples of a sentence.
\textit{Classification}: We adopt the RoBERTa model~\cite{liu2019roberta} to evaluate the results when treating the prediction of the number of views as a sequence classification task. %\newcommand{\indicator}[1]{\mathcal{I}\left(#1\right)}
We set the classes based on the number of sentiment tuples. 
%\subsection{Results}
As shown in Figure~\ref{fig:4}, the distribution of tuple counts is skewed towards the lower end, with instances containing more than seven tuples being nearly non-existent. Consequently, we limit the categories from 1 to 6 and clip instances with 7 or more tuples to 6. Additionally, to address label imbalance, we employ a weighted loss function, where the weights are set as the inverse of the frequency ratio for each category as in Equation~\eqref{eq:3}. We use the same notation as in Section~\ref{sec:2.1}, and $\indicator{}$ denotes the indicator function. This approach enables the model to effectively classify even the less represented classes. 
%We describe the detailed formula in 
%We use same notation as in Section ~\ref{sec:2.1}, and $\indicator{\cDOT}$ denotes indicator function.

\begin{equation}%\label{weighted loss}
\begin{aligned}
\boldsymbol{W_c} &= \frac{|D|}{\sum_{D}\mathcal{I}(min(\boldsymbol{|y|, 6}) = \boldsymbol{c})} \quad (c \in [1, 6]) \\
\mathcal{L}_{cls} &= -\sum_{i=1}^{|B|} \boldsymbol{W_{k_i}} \log p(\boldsymbol{k_i} | \boldsymbol{x_i}) \indicator{\boldsymbol{k_i \leq 6}}
\label{eq:3}
\end{aligned}
\end{equation}

\begin{figure}[t]
    \raggedleft
    \begin{tikzpicture}
        \begin{axis}[
            xlabel={number of sentiment tuples},
            ylabel={frequency ratio},
            xmin=0.8, xmax=7,
            ymin=0, ymax=0.8,
            xtick={1, 2, 3, 4, 5, 6, 7},
            legend pos=north east,
            xmajorgrids=true,
            ymajorgrids=true,
            grid style=dashed,
            width=0.4\textwidth,
            legend style={font=\small},
        ]
        \addplot[
            color=green,
            mark=*,
            ]
            coordinates {
            (1,0.598)(2,0.245)(3,0.106)(4,0.044)(5,0.004)(6,0.002)(7,0.001)
            };
        \addplot[
            color=red,
            mark=*,
            ]
            coordinates {
            (1,0.620)(2,0.239)(3,0.098)(4,0.037)(5,0.003)(6,0.0008)(7,0.0015)
            };
        \addplot[
            color=blue,
            mark=*,
            ]
            coordinates {
            (1,0.716)(2,0.199)(3,0.055)(4,0.018)(5,0.005)(6,0.005)(7,0.002)
            };
        \addplot[
            color=yellow,
            mark=*,
            ]
            coordinates {
            (1,0.601)(2,0.243)(3,0.105)(4,0.039)(5,0.006)(6,0.003)(7,0.002)
            };
        \legend{ASQP-Rest15, ASQP-Rest16, ACOS-Lap16, ACOS-Rest16}
        \end{axis}
    \end{tikzpicture}
    \caption{Distribution of the number of sentiment tuples. The sources are from training datasets of each task. We normalize each count by dividing it by the total number of data points. The number of tuples is clipped to 7.}
    \label{fig:4}
\end{figure}

\subsection{Effect of Element Exclusions}
\label{sec:exclude}
We analyze the impact of excluding various marker tokens, including the [O] token representing opinions, to determine which token exclusions contribute to performance improvements. Additionally, we experiment with cases where no element exclusion is performed. In this section, we have also included the second stage results to provide a detailed comparison of the overall performance.

\begin{table}[t]
\centering
\setlength{\tabcolsep}{4pt}
\small
\resizebox{0.4\textwidth}{!}{%
\begin{tabular}{l||cc|c}
\toprule
\multirow{2}{*}{\textbf{Methods}} & \multicolumn{2}{c|}{\textbf{First stage}} & {\textbf{Second stage}} \\
& \textbf{RMSE} & \textbf{Acc.} & \textbf{F1 score} \\
\midrule
Random  & 2.80  & 18.89  & -  \\
Majority  & 0.99  & 63.39  & -  \\
Classification  &  0.83   &  61.90  &  - \\
$DoT_{first}$  & 0.54  & 77.83  & \textbf{54.33}  \\
\midrule
exclude $[C]$ &  0.54   &  77.53  &  53.91  \\
exclude $[A]$  & \textbf{0.53}  & 77.77  & 53.71  \\
exclude $[S]$  &  0.54  &  77.65  &  53.55    \\
full elements & 0.55 & \textbf{78.22} & 53.94 \\
\bottomrule
\end{tabular}
}
\caption{
First stage results for each main baseline and exclusion of specific tokens. We report average RMSE loss and accuracy for first stage, and F1 score for second stage.
}
\label{tab:first}
\vspace{-3mm}
\end{table}

As in Table~\ref{tab:first}, our proposed method outperforms the other baselines and nearly predicts the actual distribution of sentiment tuples within a small margin of error. This result justifies the use of the output from the first stage in the second stage. The first stage results in Table 4 do not exhibit significant performance differences among various exclusion. However, for the second stage results, which serve as the final output of this task, we observe a significant performance difference. The performance in the second stage is generally higher when O is omitted because generating O correctly is the most difficult and crucial task in quadruple prediction~\citep{chebolu2023review}. If O is not trained in the first stage and is reused in the second stage, the model appears to focus more on learning about O compared to other elements, which already have some level of information.

\section{Computing Inference Time}
\label{sec:C}

We compare inference times based on view methods across different dataset sizes. The dataset consisted of randomly sampled test data from laptop16, with 200, 400, 600, and 800 samples. The baselines were set as static single view (T5-paraphrase) and static multi view (MvP), with the number of views for the multi view fixed at 15. 
\begin{figure}[t]
    \raggedleft
    \begin{tikzpicture}
        \begin{axis}[
            xlabel={Dataset Size},
            ylabel={Inference Time (s)},
            xmin=150, xmax=830,
            ymin=0, ymax=2850,
            xtick={200,400,600,800},
            legend pos=north west,
            xmajorgrids=true,
            ymajorgrids=true,
            grid style=dashed,
            width=0.4\textwidth,
            legend style={font=\small},
        ]
        \addplot[
            color=green,
            mark=triangle*,
            ]
            coordinates {
            (200,13.7)(400,25.7)(600,36.8)(800,46.1)
            };
        \addplot[
            color=red,
            mark=*,
            ]
            coordinates {
            (200,91.3)(400,161.72)(600,244.3)(800,322.5)
            };
        \addplot[
            color=blue,
            mark=square,
            ]
            coordinates {
            (200,974.7)(400,1393.74)(600,2040.02)(800,2721.6)
            };
        \legend{Single-view, Dynamic-view, Multi-view}
        \end{axis}
    \end{tikzpicture}
    \caption{Inference time among dataset size for each model.}
    \label{fig:3}
\end{figure}
Figure~\ref{fig:3} shows that we not only dramatically reduce inference time of utilizing multi views, but also reduce the rate of increase in inference time with respect to the number of datasets. 
On the other hand, in terms of single view, we significantly increase F1 performance while suppressing the increase in inference time and the rate of its increase. These results suggest that the efficiency of our method becomes more pronounced as the dataset size increases.

\section{Input and Target Examples for Each Stage}
\label{sec:D}

\begin{figure*}[!t]
    \centering
    \includegraphics[width=\textwidth]{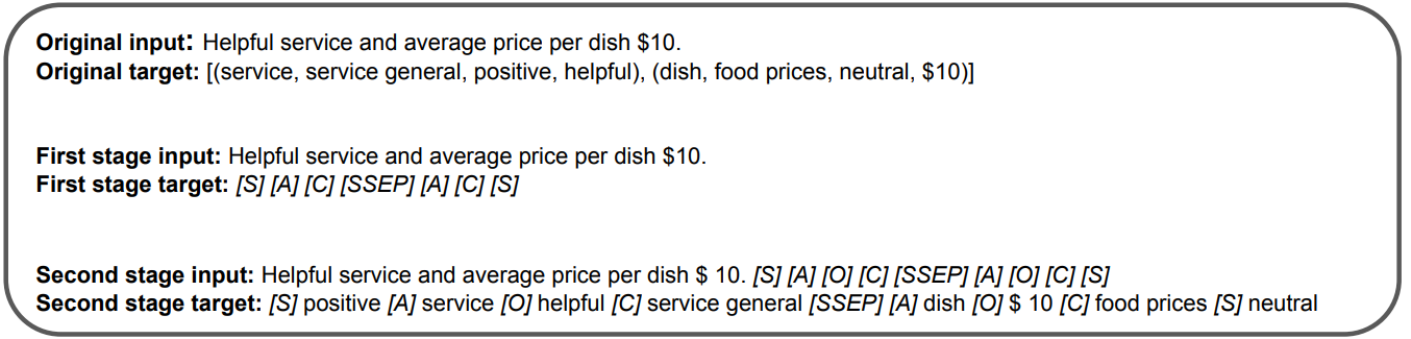}
    \caption{Examples for input and target from original dataset for both first and second stage.
    }
    \label{fig:exam}
\end{figure*}

In Figure~\ref{fig:exam}, we provide detailed examples for input and output pairs in each stage. The input sentences in the dataset are presented in a basic sentence structure, while the labels consist of lists of sentiment tuples. To preprocess this data, during the first stage, the original input sentence is kept unchanged, and the target is set as the initial order template, which consisted of a number of views corresponding to the number of sentiment tuples in the label. In the second stage, the input is processed by appending the final order template as a prompt to the original input sentence. The target is then constructed by adjusting the order of the elements within the sentiment tuples to align with the corresponding views in the order template.

\section{Analysis on Implicit Term}
\label{sec:implicit}
In Table~\ref{tab:1}, DOT suffers from predicting sentiment tuples in Rest and Clothing domains. We noted that the ACOS dataset contains a significant number of instances with implicit aspects or implicit opinions. Additionally, we discovered that the Rest and Clothing dataset are smaller in scale compared to other ACOS datasets. The scale of each dataset and the number of instances containing implicit terms are recorded in Table~\ref{tab:sample_data_ratios}. Based on these observations, we hypothesized that the size of the dataset and the distribution of implicit terms contribute to the performance degradation observed in the Rest and Clothing datasets.

\begin{table*}[t]
\centering
\setlength{\tabcolsep}{4pt}
\small
\resizebox{0.8\textwidth}{!}{%
\begin{tabular}{l|cc|cc|ccccc}
\toprule
\multirow{2}{*}{\textbf{Datasets}} & \multicolumn{2}{c|}{\textbf{ASQP}} & \multicolumn{2}{c|}{\textbf{ACOS}} & \multicolumn{5}{c}{\textbf{MEMD}}\\ 
                   & \textbf{R15} & \textbf{R16} & \textbf{Lap} & \textbf{Rest} & \textbf{M-Rest} & \textbf{M-Laptop} & \textbf{Books} & \textbf{Clothing} & \textbf{Hotel}\\
\midrule 
total samples & 834 & 1264 & 2934 & 1530 & 3622 & 2863 & 2092 & 1674 & 2481\\ 
implicit samples & 272 & 446 & 1826 & 822 & 1801 & 1751 & 1523 & 1083 & 1278\\ 
implicit sample \% & 32.6 & 35.3 & 62.2 & 53.7 & 49.7 & 61.2 & 72.8 & 64.7 & 51.5\\ 
\bottomrule
\end{tabular}
}
\caption{The size of each dataset and the number of samples containing implicit terms. For ease of comparison, We also provide the percentage of samples with implicit terms relative to the total number of samples. It is evident that the implicit term ratio in the ACOS dataset is higher compared to that in the ASQP dataset.}
\label{tab:sample_data_ratios}
\end{table*}

\begin{table}[h]
\centering
\setlength{\tabcolsep}{4pt}
\small
\resizebox{0.45\textwidth}{!}{%
\begin{tabular}{c|c|c|c|c}
\toprule
\textbf{Methods} & \textbf{Rest} & \textbf{M-Rest ¼} & \textbf{M-Rest ½} & \textbf{M-Rest full} \\ 
\midrule
Paraphrase & \textbf{50.06} & \textbf{40.26} & 47.77 & 49.09 \\ 
DOT & 44.84 & 35.74 & \textbf{47.81} & \textbf{49.49} \\ 
\bottomrule
\end{tabular}
}
\caption{F1 scores only for samples containing implicit terms. We report the performance in Rest dataset and the performance trends across different dataset scales.}
\label{tab:comparison_methods}
\end{table}

As shown in Table~\ref{tab:comparison_methods}, it is evident that the F1 score for instances containing implicit terms in the Rest dataset is significantly lower compared to using the paraphrase method. Additionally, we observed a performance degradation when training on a randomly selected quarter of the M-Rest dataset. However, as the amount of training data from the M-Rest dataset increased, the performance on implicit terms improved, eventually surpassing the F1 score of the paraphrase method in the full M-Rest dataset. This result demonstrates that the small size of the dataset with a high proportion of implicit terms is the primary cause of the performance degradation in the Rest and Clothing dataset. It also suggests that the performance is likely to improve as the dataset size increases.

\section{Additional Analysis}
\label{sec:Additional}
In this section, we conduct an in-depth analysis of various aspects of our model. For a comprehensive evaluation, we use Paraphrase and MvP as baselines, running identical experiments for comparison. We assess performance across multiple tasks using several benchmarks, including R15, R16, Lap, Rest, and M-Rest.

\paragraph{Different Backbone Model} We conduct the experiment using different encoder-decoder based model, BART~\cite{lewis2019bart} as backbone model. We utilize BART with the same hyperparameters and data processing techniques applied to the T5 model for three methods including ours. However, as in Table~\ref{tab:Bart}, we observe a noticeable decline in overall F1-scores for all models, primarily due to insufficient hyperparameter tuning compared to T5. Nevertheless, as shown in the results, our method still outperforms the baseline models with BART, suggesting that its effectiveness is not highly dependent on the choice of backbone model.
\begin{table}[h]
\centering
\small
\resizebox{0.45\textwidth}{!}{
\begin{tabular}{l|cc|ccc}
\toprule
\multirow{2}{*}{\textbf{Methods}} 
& \multicolumn{2}{c|}{\textbf{ASQP}} 
& \multicolumn{3}{c}{\textbf{ACOS}} \\ 
& \textbf{R15} & \textbf{R16} & \textbf{Lap} & \textbf{Rest} & \textbf{M-Rest}\\
\midrule 
Paraphrase & 31.77 & 38.15 & 30.98 & 36.65 & 35.74 \\ 
MvP        & 33.48 & 41.01 & 32.57 & \textbf{40.40} & 40.30 \\ 
DOT        & \textbf{35.98} & \textbf{41.73} & \textbf{33.12} & 39.61 & \textbf{40.91} \\ 
\bottomrule
\end{tabular}
}
\caption{F1 score on benchmark datasets using BART as the backbone model.}
\label{tab:Bart}
\end{table}

\paragraph{Complex Sentences} As mentioned in Appendix~\ref{sec:case_study}, processing long and complex contexts is a well-known challenge, and our model performs similarly to others in this regard. We define complex sentences as those containing more than three sentiment tuples, exceeding 22 words in length, or having a Flesch-Kincaid Grade Level~\cite{solnyshkina2017evaluating} of 9 or higher, representing the top 20\% for each criterion. We sample these complex sentences and evaluate the F1 scores for these samples. As shown in the Table~\ref{tab:Complex}, performance degradation in complex sentences is a common issue across all models. We attribute the larger performance drop in our model compared to MvP to the fact that it uses fewer views, which limits its capacity to thoroughly analyze complex sentences. Nevertheless, our model's final Complex F1 score remains close to that of MvP and surpasses that of the Paraphrase model.
\begin{table}[h]
\centering
\small
\resizebox{0.45\textwidth}{!}{
\begin{tabular}{l|cc|ccc}
\toprule
\multirow{2}{*}{\textbf{Methods}} 
& \multicolumn{2}{c|}{\textbf{ASQP}} 
& \multicolumn{3}{c}{\textbf{ACOS}} \\ 
& \textbf{R15} & \textbf{R16} & \textbf{Lap} & \textbf{Rest} & \textbf{M-Rest}\\
\midrule 
Paraphrase & 44.94 & 55.73 & 37.15 & \textbf{56.12} & 54.37 \\ 
MvP        & 46.71 & \textbf{58.00} & 37.68 & 56.06 & \textbf{58.23} \\ 
DOT        & \textbf{46.93} & 57.70 & \textbf{38.57} & 54.16 & 57.72 \\ 
\bottomrule
\end{tabular}
}
\caption{F1 scores evaluated on complex samples only.}
\label{tab:Complex}
\end{table}

\paragraph{Training Complexity} Our method may appear complex due to the numerous components that require training. However, since our method involves simply training the T5 model twice without complex optimization procedures, the overall training time is not significantly longer than that of other models.
As shown in the Table ~\ref{tab:time}, even though our model uses 30 and 40 epochs for two stages of training—more than the 20 epochs used in MvP—the total training time remains much shorter than that of MvP. In terms of memory usage, only two T5 models are allocated in memory, so the memory consumption does not increase exponentially compared to existing models.
\begin{table}[h]
\centering
\small
\resizebox{0.45\textwidth}{!}{
\begin{tabular}{l|cc|ccc}
\toprule
\multirow{2}{*}{\textbf{Methods}} 
& \multicolumn{2}{c|}{\textbf{ASQP}} 
& \multicolumn{3}{c}{\textbf{ACOS}} \\ 
& \textbf{R15} & \textbf{R16} & \textbf{Lap} & \textbf{Rest} & \textbf{M-Rest}\\
\midrule 
Paraphrase & 212.83 & 314.18 & 652.31 & 349.79 & 815.48 \\ 
MvP        & 3883.74 & 5008.84 & 11006.07 & 6169.02 & 14634.71 \\ 
DOT        & 1161.73 & 1648.61 & 3310.63 & 1814.41 & 4157.93 \\ 
\bottomrule
\end{tabular}
}
\caption{Training duration for each benchmark. }
\label{tab:time}
\end{table}

\paragraph{Standard Deviation} We conduct experiments using five different random seeds and calculate the standard deviation of the outcomes. Results are reported in Table~\ref{tab:sd}. Our findings indicate that our model exhibits a higher overall standard deviation compared to other baselines. This can be attributed to the structure of the method, where an error at one stage is likely to propagate and accumulate. However, it is important to note that the absolute value of the standard deviation is not significantly large. In fact, the higher variation suggests that the model may possess greater potential to achieve stronger performance.
\begin{table}[h]
\centering
\small
\resizebox{0.45\textwidth}{!}{
\begin{tabular}{l|cc|ccc}
\toprule
\multirow{2}{*}{\textbf{Methods}} 
& \multicolumn{2}{c|}{\textbf{ASQP}} 
& \multicolumn{3}{c}{\textbf{ACOS}} \\ 
& \textbf{R15} & \textbf{R16} & \textbf{Lap} & \textbf{Rest} & \textbf{M-Rest}\\
\midrule 
Paraphrase & $\pm$ 0.44 & $\pm$ 0.64 & $\pm$ 0.26 & $\pm$ 0.68 & $\pm$ 0.38 \\ 
MvP        & $\pm$ 0.54 & $\pm$ 0.29 & $\pm$ 0.48 & $\pm$ 0.72 & $\pm$ 0.48 \\ 
DOT        & $\pm$ 0.74 & $\pm$ 0.85 & $\pm$ 1.01 & $\pm$ 0.76 & $\pm$ 0.42 \\ 
\bottomrule
\end{tabular}
}
\caption{Standard deviation of outcomes in Table~\ref{tab:1}.}
\label{tab:sd}
\end{table}

\section{Detailed Setups for LLM Experiments}
\label{sec:llm}
\begin{figure*}[h]
    \centering
\begin{lstlisting}[ 
        basicstyle=\small\ttfamily, 
        label=lst:list1,
        breaklines=true,
        frame=single,
        stepnumber=1,
        texcl=true,
        escapeinside={(*@}{@*)},
        %moretexcs={textbf, textit}
    ]
According to the following sentiment elements definition: 

- The 'aspect term' refers to a specific feature, attribute, or aspect of a product or service that a user may express an opinion about, the aspect term might be 'null' for implicit aspect.
- The 'opinion term' refers to the sentiment or attitude expressed by a user towards a particular aspect or feature of a product or service, the aspect term might be 'null' for implicit opinion.
- The 'aspect category' refers to the category that aspect belongs to, and the available categories includes: {(*@\textbf{dataset specific categories}@*)}.
- The 'sentiment polarity' refers to the degree of positivity, negativity or neutrality expressed in the opinion towards a particular aspect or feature of a product or service, and the available polarities inlcudes: 'positive', 'negative' and 'neutral'.

Recognize all sentiment elements with their corresponding aspect terms, aspect categories, opinion terms and sentiment polarity in the following text with the format of [('aspect term', 'aspect category', 'sentiment polarity', 'opinion term'), ...]: 
\end{lstlisting}
    \includegraphics[width=1\textwidth]{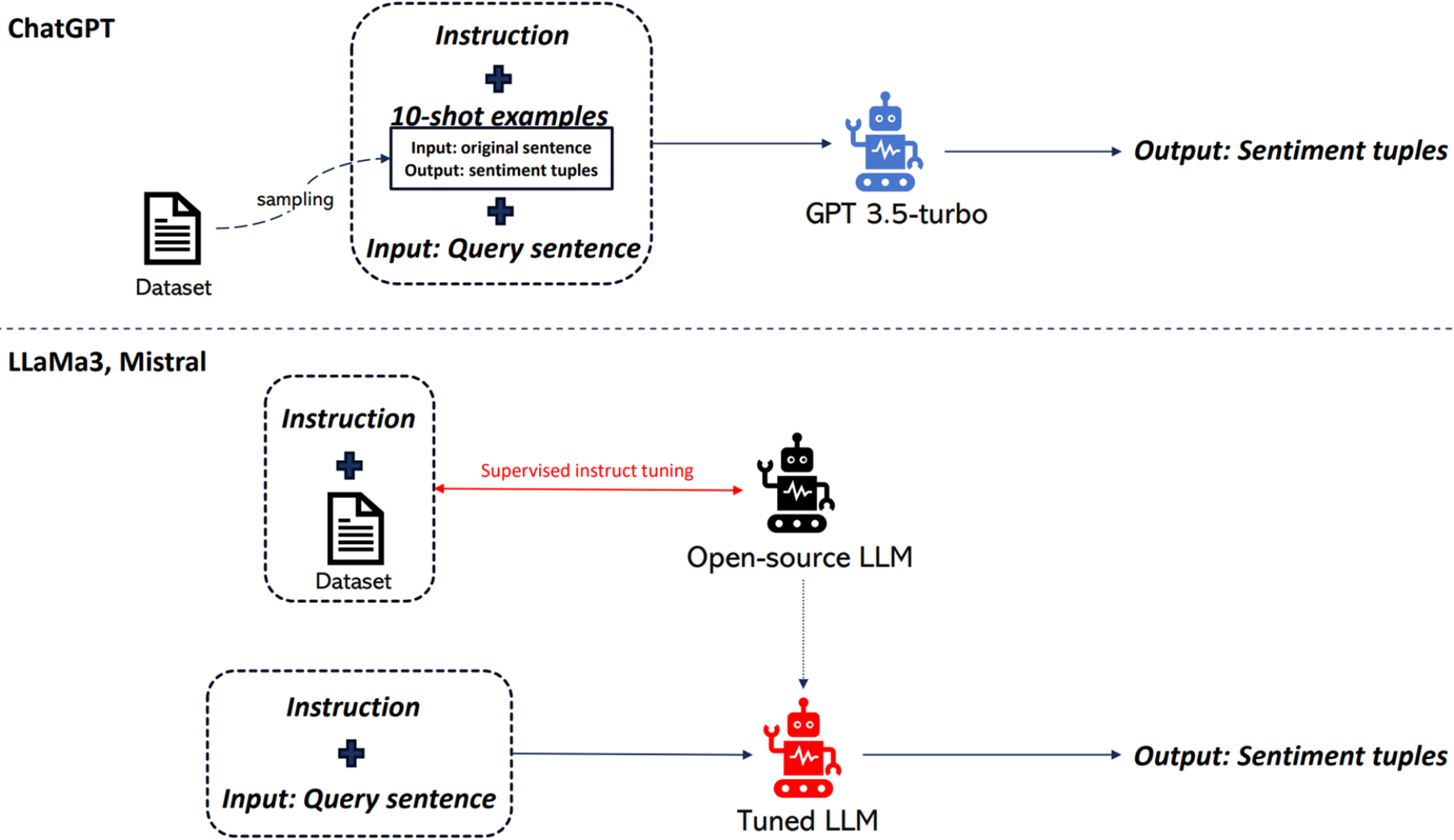}
    \caption{
    Instruction format for two LLM frameworks. We utilize in-context learning for GPT-3.5-turbo inference, and instruction-tuning for LLaMa-3.1 and Mistral inference respectively.
    }
    \label{fig:7}
\end{figure*}

As in Table~\ref{tab:1}, we perform the ABSA task using the GPT-4o, LLaMa-3.1-8B, and Mistral-7B models, compairing the results with our DOT model. For the GPT model, we utilize in-context learning~\cite{brown2020language}. We randomly sample 10 instances and combine them with instruction format, and add it as a prompt. For the other three open-source LLMs, we employ instruction tuning~\cite{wei2021finetuned} with the training dataset for fine-tuning, using the same instructions as in GPT prompts. To ensure stable model training during fine-tuning, we utilize the LoRa~\cite{hu2021lora}. We present the specific prompts and framework in Figure~\ref{fig:7}.

%This is an appendix.
\end{document}